\documentclass[a4paper,10pt,conference]{IEEEtran}

\usepackage[dvips]{graphics}
\usepackage{hyperref}
\usepackage{graphicx}

\usepackage[latin1]{inputenc}  
\usepackage[T1]{fontenc}       

\usepackage{fancybox,fancyhdr}

\begin{document}

 \thispagestyle{fancy}
 \lhead{Accepted paper to "Supervisory Control in Critical Systems Management" Workshop at Human Centered Processes 2008 Conference (HCP-2008), June 8-12, 2008, Delft, The Netherlands. }
\pagestyle{empty}

\title{Collaborative model of interaction and\\
             Unmanned Vehicle Systems' interface}

\author{

\IEEEauthorblockN{Sylvie Saget}
\IEEEauthorblockN{Fran\c{c}ois Legras}
\IEEEauthorblockN{Gilles Coppin}

\IEEEauthorblockA{TELECOM Institute, TELECOM Bretagne, LUSSI Department\\
Email: \{sylvie.saget,francois.legras,gilles.coppin\}@telecom-bretagne.eu}
}

\maketitle

\begin{abstract}
\boldmath
\indent The interface for the next generation of Unmanned Vehicle Systems should be an interface with multi-modal displays
and input controls. Then, the role of the interface will not be restricted to be a support of the interactions between the
ground operator and vehicles. Interface must \textit{take part} in the interaction management too.\newline
\indent In this paper, we show that recent works in pragmatics and philosophy \cite{SG06} provide a suitable theoretical
framework for the next generation of UV System's interface. We concentrate on two main aspects of the collaborative model
of interaction based on acceptance: multi-strategy approach for communicative act generation and interpretation and communicative alignment.
\end{abstract}


\section*{Introduction}

\indent At the moment, most Unmanned Vehicle (UV) Systems are single vehicle systems whose control mode is teleoperation. Several ground operators are needed in order to operate a vehicle. Besides, vehicles have limited autonomous capabilities. Consequently, controlling vehicle is such a hard task that it may lead to an untractable cognitive
load for the ground operator \cite{MGKH01}. In order to make this task more feasible and in order to reduce the cost of UV Systems in term of human resource, several areas of reflection are explored:
\begin{itemize}
    \item drifting from UV system with a single vehicle to UV system with multiple vehicles \cite{Joh03},
    \item increasing vehicle's autonomy \cite{DW01}.
\end{itemize}
\indent As a result, control mode will shift to a more flexible control mode such as control/supervision in the next
generation of UV Systems. Moreover, the role of the operator will shift to controlling/supervising a \textit{system} of
several cooperating UVs performing a joint mission \textit{i.e.} a Multi-Agent System (MAS) \cite{LC07}. \newline
\newline
\indent In the same time, current works aim at enhancing the flexibility and the naturalness of interface rather than only
improving the mission's realization and control. In particular, human-centered approaches introduce new modalities
(gesture, spoken or written language, haptic display, etc.), \cite{MGKH01,GNBWSC02}. The interface for the next generation
of UV Systems should be an interface with multi-modal displays and input controls. Actually, multi-modal displays aim at
making up for the "sensory isolation" of ground operator, as well as reducing cognitive and perceptual demands
\cite{GNBWSC02}. This is especially important considering the high visual demand of such interface. Moreover, multi-modal
controls aims at reducing cognitive workload as well as at making operator's control more efficient \cite{WDCB05}. For
example, a data entry function based on vocal keyword recognition may require a single vocal utterance in the next generation of interface, while it
may require over twenty separate manual actions in the current generation of interface. Furthermore, human control of MAS, such as control-by-policy or playbook, may require highly flexible and less constrained language interaction \cite{LC07}.
\newline
\indent Then, the role of the interface is not restricted to be a support of interactions between the ground operator and vehicles. The interface must also \textit{transcribe} the communicative information in the suitable presentation mode for each dialog partner. Besides, future interfaces must also provide tools in order to make the interaction management easier. For example, the management of interactions with several vehicles by the ground operator at the same time is quite complex. Interfaces must \textit{take part} in the interaction management too. Actually, non-understandings are frequent in "natural" multi-modal interaction. An utterance of the ground operator may not have been perceived because of background noise, an utterance may be not-understood because of an unknown word, a gesture or an utterance may be ambiguous or incoherent, etc. A control input can be transmitted to vehicles by the interface only if this control input has been understood.
Thus, the interface has to manage such non-understandings. Inversely, the interface has to manage ground operator's attempts for clarification of non-understood multi-modal display.\newline
\newline
\indent The collaborative nature of interaction (or dialogue) have been brought into the forefront by research in pragmatics since mid-90s \cite{Cla96}. Basing an interface's interaction management on such a model gives the interface and its users the capacity to interactively refine their understanding until a point of intelligibility is reached. Thus, such interface manages non-understandings\footnote{Non-understanding is commonly set apart misunderstanding. In a misunderstanding, the addressee succeeds in communicative act's interpretation, whereas in a non-understanding he fails. But, in a misunderstanding, addressee's interpretation is incorrect. For example, mishearing may lead to misunderstanding. Misunderstandings are considered here as the only kind of "communicative errors" (c.f. section \ref{TradCollab}). Thus, they are handled by a recovery process, which is not supported by the interaction model.}. This approach have been used within the WITAS dialog system \cite{LGCP02}.\newline
\indent In this paper, we show that recent works in pragmatics and philosophy \cite{SG06}  provide a suitable theoretical framework for the next generation of UV System's interface. We concentrate on two main aspects of the collaborative model of interaction based on acceptance: multi-strategy approach for generation and interpretation of communicative acts and communicative alignment.

\section{Preliminary distinctions}

    \subsection{Task level versus Interaction level}

\indent While using an UV System's interface, the ground operator is at least engaged within two activities: mission control/supervision and interaction. This is the general case of all goal-oriented interaction (or dialogue):
\begin{quote}
"Dialogues, therefore, divide into two planes of activity \cite{Cla96}. On one plane, people create dialogue in service of the basic joint activities they are engaged in-making dinner, dealing with the emergency, operating the ship. On a second plane, they manage the dialogue itself-deciding who speaks when, establishing that an utterance has been understood, etc. These two planes are not independent, for problems in the dialogue may have their source in the joint activity the dialogue is in service of, and vice versa. Still, in this view, basic joint activities are primary, and dialogue is created to manage them." \cite{BC03}.
\end{quote}

Interaction is defined by dialog partner's goals to understand each other, in other words to reach a certain degree of intelligibility, \textit{sufficient for the current purpose}.\newline
 \newline
\indent The crucial points here are that :
\begin{enumerate}
    \item perfect understanding is not required, the level of understanding required is directed by the basic activity
    (\textit{i.e.} the mission) and the situational context (\textit{i.e.} time pressure for example);
    \item as ground operator's cognitive load is "divided" between the cognitive load induced by each activity, the interaction's complexity must vary depending on the complexity involved by the mission \cite{MGKH01}. For example, as time pressure rises, the cognitive load induced by the mission increases. The cognitive load required by the interaction must decrease in order to carry through the mission.
\end{enumerate}
\indent All in all, a model of interaction dedicated to UV System's interface has to support multi-strategy methods for communicative acts generation and interpretation.\newline
\indent However, one may bring together generation and interpretation methods in two main types: methods following pragmatics fundament (\textit{i.e.} interaction model), such as the sincerity hypothesis or the maxim of manner \cite{Gri75}, and methods which do not. The first type aims at reaching high quality of understanding but are complex. The second type aims at efficiency but quality of understanding is not ensured. Each kind of methods is mono-strategic or support a little set of possible strategies. Existing methods are interpretation based on keyword recognition \cite{SEM07}, statistical methods based on heuristics \cite{Wan01}, more pragmatics-based approach \cite{GS07}, etc.\newline
In this paper (section \ref{OurModel}), we present an interaction model which is coherent with each type of method. Thus, an interaction manager based on such a model can support multi-strategy methods of communicative acts generation and interpretation.
\label{Multi-strat}

    \subsection{Interaction model versus Interaction management}

\indent For methodological reasons, the distinction between interaction model (dialogue model) and interaction manager
(dialogue manager) has to be made clear \cite{XXHX02}. An \emph{interaction model} is a theoretical model which aims at providing a general theory of interaction. An \emph{interaction manager} is  a component of an interface, as shown by the Fig. 1.\newline
\indent An interface perceives events such as control input (communicative act). After the perception of an event,
the interface interprets it. The role of the interaction manager is to decide which is the suitable reaction. Following a
control input from the ground operator, a possible reaction is to transmit the command to the proper vehicle, if the
control input has been understood. Another possible reaction is to ask for clarification to the ground operator if the
control input is ambiguous. Inversely, following an ambiguous display, requests for clarifications by the ground operator
have to be supported by the interaction manager.
\begin{figure}[!ht]
        \begin{center}
       \includegraphics*[width=10.53cm,height=4.35cm]{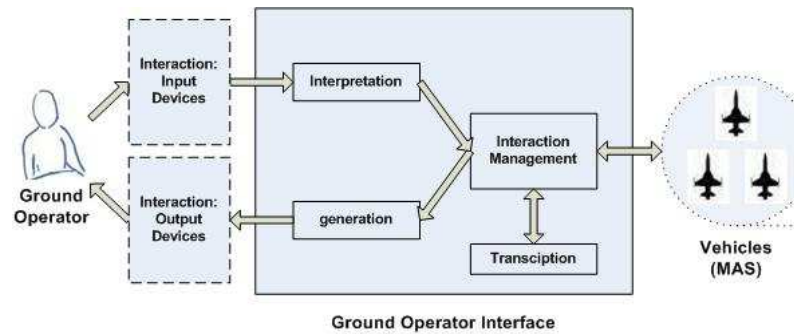}
       \caption{Architecture of an interface.}
       \label{FigComposantSystDial}
    \end{center}
\end{figure}

\indent There are several kind of technological tools dedicated to interaction management. Each kind of technological tool is based on a model of interaction, as shown by Fig. 2.
    \begin{figure}[!ht]
    \begin{center}
    \begin{tabular}{|c|c|}
      \hline
      \small{\textbf{Manager}}  &  \small{\textbf{Model}} \\
      \hline
      Dialog Grammar           & Adjacent Pairs \cite{SSJ78}\\
      \hline
      Plan-based manager       & Speech Acts Theory \cite{Sea69} \\
      \hline
      Agent-based manager       & Cognitive models \cite{CL90,Sad94b} \\
      \hline
    \end{tabular}
       \caption{Interaction managers and corresponding interaction models.}
    \end{center}
\end{figure}

\indent The choice of an interaction manager is mainly based on the task (mission) underlying the interaction \cite{ABD01}.
However, each kind of interaction model captures a particular aspect of interaction and may have consequences on each kind
of interaction manager. Cognitive models of interaction aim, for instance, at defining a symbolic and explanatory model of
interaction, whereas Adjacent Pairs provide a descriptive model of interaction. Cognitive models may be considered as a
logical reformulation of plan-based models. Cognitive models integrate, in more, a precise formalization of dialog partners'
mental states (their beliefs, choices (or desires) and intentions), of the rational balance which relates mental attitudes between them and relates mental attitudes with agents' acts.

\section{Collaborative model of interaction}

\indent Basing interaction management on a collaborative model of interaction gives the interface the ability to manage non-understandings, as shown in the first part of this section.\newline
\indent A formal collaborative model of interaction is generally based on a psycholinguistic model of interaction. However, existing psycholinguistic models of interaction do not support  multi-strategy approach for communicative act generation and interpretation. We propose to base interaction management, for the next generation of UV Systems, on a formal interaction model supporting such a multi-strategy approach. This formal model mixes and enhances the two main and complementary psycholinguistic models of interaction. The second part of this section introduces these two psycholinguistic models of interaction.

    \subsection{Traditional view vs. Collaborative view of interaction}

\label{TradCollab}

\indent The traditional view of interaction \cite{Gri75,SSJ78,Sea69,CL90,Sad94b} defines it as an \textit{unidirectional}
process resulting from two individual activities: the generation of a communicative act by the "speaker" and the
understanding and interpretation of this communicative act by the addressee. Interaction's success is warranted by the
cooperative attitude of the "speaker" (his sincerity, his relevance, etc.). Consequently, the production of a suitable
communicative act is concentrated on a single exchange and a single agent. The complexity (\textit{i.e.} the cognitive
load) of such a process is high because of the necessary restrictive hypothesis \cite{GP04}. Moreover, the set of
possible strategies to produce and understand a communicative act is highly limited. Besides, the addressee having a
passive role, positive feedbacks such as "Okay", "Mhm", "Uhuh", head nodes, etc., signalling successful understandings,
are not necessary. Finally, non-understandings are regarded as "communicative errors" which have to be handled by extra complex mechanisms. \newline
\newline
\indent In contrast with the traditional view, collaborative model of interaction defines it as a \textit{bidirectional} process resulting from a single social activity. Interaction is considered as a collaborative activity characterized by the goal of reaching mutual understanding, shared by dialog partners. Mutual understanding is reached through interpretation's negotiation. That is an interactive refinement of understanding until a sufficient point of intelligibility is reached, illustrated by the example shown in Fig. 3.
\begin{figure}[!ht]
    \begin{itemize}
            \item[] User: Go to the building.
            \item[] System: Which building do you mean?\
            \item[] System: I can see a blue car at the tower.
            \item[] System: It is driving on Creek Lane.
            \item[] System: Warning my fuel is low.
            \item[] User: I mean the school.
       \caption{An example of interpretation negotiation, \cite{LGGBHP03}.}
    \end{itemize}
\end{figure}

Consequently, the production of a suitable communicative act can be divided between several exchanges and between all
dialog partners. The complexity of such process must be less complex than in the traditional view of interaction
\cite{GP04}. Besides, the addressee has an active role, explicit and implicit feedbacks are required in order to publicly signal
successful understandings. Finally, non-understandings are here regarded as "the normal case", so their management is captured by collaborative model of interaction

    \subsection{Two complementary models}

            \subsubsection{Clark's Intentional model}

\indent Most of formal collaborative models of interaction are based on the psycholinguist H. H. Clark's work \cite{Cla96,BC96}. His work highlights the collaborative nature of interaction, its realization through a negotiation process, its success warranted by the use of the common ground (\textit{i.e.} mutual beliefs) among dialog partners, conceptual pacts (\textit{i.e.} temporary, partner-specific alignment among dialog partners on the description chosen for a particular object).\newline
\indent Basing interaction management on this model is interesting because:
\begin{enumerate}
    \item Designing interaction as a collaborative process enhances mixed-initiative interaction.
    \item Non-understandings are interactively managed, thus interface's robustness and flexibility are enhanced.
    \item Positive and negative signals of understandings are consistently required, as part of the negotiation process.
    \newline
\end{enumerate}
\indent However, there are several limitations against this model \cite{SG06}:
\begin{enumerate}
    \item The systematic use of common ground leads to mono-strategic and complex generation and interpretation of communicative acts. In Human-Human interactions, dialog partners rely on different strategies. The complexity of the strategy vary depending on the context, depending on time pressure for example.
    \item Considering common ground as a set of mutual beliefs leads to computational limitations and paradoxes, as human beings tends to have selfish and self-deceptive attitudes.
    \newline
\end{enumerate}
\indent To sum up, this model is suitable for modeling non-understandings management through interpretation negotiation. Nevertheless, interpretation negotiation, as defined in this model, is too restrictive. This is due to systematic use of common ground and defining common ground as a set of mutual beliefs, \textit{i.e.} a stronger definition of the sincerity hypothesis.
\newline
            \subsubsection{The Interactive Alignment Model}
\indent Another model of the collaborative nature of interaction has been proposed by  M. J. Pickering and S. Garrod \cite{PG04}: the Interactive Alignment Model (IAM). IAM claims that dialog partners become aligned at several linguistics aspects. In the particular case of spoken dialog, there is an alignment, for example, of the situation model, of the lexical and the syntactic levels, even of clarity of articulation, of accent and of speech rate. \newline
\indent For example, syntactic alignment is frequent in question-answer, such as in Fig. 4.
\begin{figure}[!ht]
    \begin{itemize}
            \item[] User: Is there a vehicle near the hospital?
            \item[] System: Yes, there are three vehicles near the hospital.
       \caption{An example of syntactic alignment.}
    \end{itemize}
\end{figure}

Reference alignment corresponds to the notion of "conceptual pacts" in Clark's model: an alignment on the description chosen to refer to a particular object during interaction, as shown in Fig. 5.
\begin{figure}[!ht]
    \begin{itemize}
            \item[] User: Keep watch at \underline{the big zone near the hospital}.
            \item[] $\vdots$
            \item[] User: There are intruders in \underline{the zone near the hospital}.
       \caption{An example of reference alignment.}
    \end{itemize}
\end{figure}

\indent These alignments results from automatic processes based on priming. Priming consists in reusing the result of a
preceding cognitive process, such as perception or action execution, in a following cognitive process. In the particular
case of interaction, priming consists in reusing words or syntactic constructions recently understood or generated. As an
automatic process, priming does not induce any cognitive load. Besides, these alignments facilitate communicative act
generation and interpretation, as well as facilitate social relationship (confidence, rapport, etc.),
\cite{LJCC03}.\newline
\newline
\indent To sum up, this model is suitable for enhancing communicative act generation and interpretation. It allows reusing
results of preceding successful interactions for the treatment of following communicative acts. Such results are part of
the common ground among dialog partners, \textit{i.e.} co-construction of "interactive" tools during interaction. IAM is viewed here as a complementary model of Clark's work. That is, each model provides an alternative strategy which can be used to generate or interpret a particular communicative act. In addition, negotiation interpretation, as described in Clark's model, manages non-understandings.
    \subsection{Collaborative interaction model based on acceptance}
       \label{OurModel}

\indent S. Garrod and M. J. Pickering claim that considering interaction as a collaborative activity must lead to avoid or to modify fundamental hypothesis responsible of several limitations \cite{GP04}. Generally speaking, dialog partners are supposed to be rational while interacting. Their rationality is partly defined by their sincerity, \textit{i.e.} they have to use (mutually) true statements in order to be understood. This sincerity hypothesis highly limits the set of possible strategies for communicative acts generation and interpretation. Thus, selfish or self-deceptive attitudes are considered as being irrational, automatic processes such as priming are not allowed, etc.\newline
\indent In preceding works, the incoherence of the systematic use of the sincerity hypothesis has been demonstrated \cite{SG06,SG07}. In fact, interaction is a goal-oriented process which aims here at transmitting informations and control orders. A particular communicative act aims at contributing to:
\newline
\begin{enumerate}
    \item enabling the addressee to interpret it,
    \item ensuring the correctness of his interpretation,
    \item contributing to mission's realization. Thus, its generation and interpretation has to be more or less efficient depending on the current task and situational context (ex. time pressure), cf. section \ref{Multi-strat}.
    \newline
\end{enumerate}
\indent The problem with the sincerity hypothesis is not that true statement can not enable to reach these goals. The problem is that there is a confusion between what is the aim of the interaction and what is the suitable strategy to use. Distinguishing these two aspects  avoid to impose a particular and single strategy.\newline
\newline
\indent In order to introduce the distinction in a collaborative model of interaction, the philosophical notion of acceptance is used \cite{SG06,SG07}. Thus, the suitable type of interaction model is cognitive model. Acceptance is the contextual mental attitude underlying a goal-oriented activity, whereas belief is the contextual mental attitude underlying a truth-oriented activity \cite{SG07}.
\newline
\newline
 $Acc_{i}(\varphi,\phi)$ stands for \textit{"the agent $i$ accepts $\varphi$ in order to bring about $\phi$"}.
\newline
\newline
Here, $\phi$ is the complex goal defined in the preceding paragraph. $\varphi$ is an association between an \textit{interactive tool} \textit{IT} (a gesture, a multi-modal display, an utterance, etc.) and the intended meaning \textit{IM} (a particular object, an order, an information, etc.):
\newline
\newline
 $\phi = communicate\_by(IM,IT)$ stands for \textit{"using $IT$ to communicate $IM$"}.
\newline
\newline
Generation is viewed as choosing an interactive tool knowing the intended meaning to convey. Interpretation is viewed as
identifying the intended meaning knowing the interactive tool. Such definitions do not set the strategy to use. Thus, all possible strategies can be considered:
\newline
\begin{itemize}
    \item priming,
    \item selfish attitude: considering solely their own belief,
    \item cooperative attitude: considering solely the addressee's beliefs or knowledge,
    \item basing interpretation on keywords recognition,
    \item etc.
    \newline
\end{itemize}
The proper strategy depends on the task, time pressure, interaction's history \textit{i.e.} depends on existing conceptual pacts, etc.\newline
\newline
\indent Concerning the interaction manager, the interaction model defines interpretation as a reactive process within a cognitive model of interaction. Following a communicative act and its interpretation, the addressee (i.e. the interface or the ground operator) is obliged to react by:
\newline
\begin{itemize}
    \item signalling his understanding through an implicit or explicit positive feedback,
    \item requesting a refinement (\textit{i.e} a clarification) of a non-understood IT or asking for a "recasting",
    \item proposing a refinement or a "recasting",
    \item postponing his reaction because of a top-priority goal to bring about.
    \newline
\end{itemize}
This is a social law, closed to the notion of negotiation protocol, which models interpretation negotiation handling
non-understanding. Based on H.H. Clark's work, this social law provides different ways of reacting following a
non-understanding. Thus, the model of interaction presented here provide multi-strategy approach for communicative act's
generation and interpretation, as well as for interaction management.

\section*{Conclusion}

Interface of the next generation of UV Systems must support multi-strategy approach of communicative act generation and interpretation. Moreover, the interface has to take part to the interaction management through non-understanding handling in particular. Our goal is to provide a suitable theoretical framework for future interaction managers. We present a collaborative model of interaction mixing and enhancing the two main psychological collaborative of interaction.\newline
\newline
\indent Further studies will hold on extending and applying our collaborative model of interaction to the particular case of  topological and tactical references used in UV Systems. First at all, we will focus on analyzing and modeling strategic choices and on defining a suitable representation of the "interactive tool".

\nocite{*}
\bibliographystyle{IEEEtran}
\bibliography{BibArticle_SCCSM-08_VF}

\end{document}